\documentclass[]{style/ceurart}
\usepackage{listings}
\begin{document}

\copyrightyear{2026}
\copyrightclause{Copyright for this paper by its authors.
  Use permitted under Creative Commons License Attribution 4.0
  International (CC BY 4.0).}

\conference{CLEF 2026: Conference and Labs of the Evaluation Forum, September 21-24, 2026, Jena, Germany}

\title{Decoupled Pipeline with Proposal Reranking and Score Fusion for Positive-Unlabeled Marine Species Detection}

\author[1]{Robert James Brock}[
    orcid=0009-0003-8059-3335,
    email=rbrock8@gatech.edu,
]
\cormark[1]

\author[1]{Sebastian Maximilian Krupa}[
    orcid=0009-0008-7763-4470,
    email=skrupa3@gatech.edu,
]
\cormark[1]

\author[1]{Jason Kahei Tam}[
    orcid=0009-0005-0853-0414,
    email=jtam30@gatech.edu,
]
\cormark[1]

\address[1]{Georgia Institute of Technology, North Ave NW, Atlanta, GA 30332}
\cortext[1]{Corresponding author.}

\begin{abstract}
The FathomNetCLEF 2026 competition combines underwater object detection and fine-grained marine species classification under a positive-unlabeled evaluation setting. The provided training labels are sparse, while the hidden test set is out-of-distribution relative to the training imagery, creating both annotation incompleteness and source-shift challenges. We describe DS@GT ARC's multi-stage system developed for this setting while keeping model training restricted to the data provided by the competition. The final private-leaderboard model uses a frozen Megalodon YOLOv8x detector as a class-agnostic proposal generator, combines global and tiled inference with tile-edge filtering, classifies expanded proposal crops with a LoRA-finetuned DINOv3 ViT-H classifier, and ranks predictions using weighted geometric fusion of detector and classifier confidence. This system placed 12th out of 102 teams. A closely related variant added a locally trained TTN-inspired validity head as a light reranking signal, improving public-leaderboard and proxy-evaluation performance but slightly reducing private-leaderboard performance. Across experiments, the strongest lesson was that train-derived validation and detector-only metrics were not reliable enough for model selection. Instead, we used proxy datasets only for validation and comparison, and combined those signals with leaderboard feedback and targeted ablations. These experiments showed that reserving proposal recall, avoiding over-aggressive filtering, and improving downstream ranking were more effective than fine-tuning the detector or directly training on noisy pseudo-labels. Code: https://github.com/dsgt-arc/fathomnetclef-2026. 
    
\end{abstract}


\maketitle

\section{Introduction}

Marine species identification from underwater imagery presents a unique challenge in computer vision due to the prevalence of partially labeled data, where annotators label only the species within their expertise, leaving other organisms unmarked \cite{katija2022fathomnetglobalimagedatabase}. This challenge is further compounded by degraded image quality inherent to aquatic environments and the vast ecological diversity of a realm covering over 70$\%$ of the Earth’s surface \cite{smithsonian_ocean_big_ocean}. This paper documents our experiments and presents our best approach for the FathomNetCLEF 2026 \cite{fathomnet2026} competition @ CVPR-FGVC \& LifeCLEF 2026 \cite{lifeclef2026}. 

Past editions of the competition focused on object detection (2024) \cite{fathomnet2024}, and taxonomic classification (2025) \cite{fathomnet2025} separately. The 2026 edition merges these tasks, focusing on positive-unlabeled (PU) learning for object detection and classification. Specifically, the competition evaluates the top-100 predictions on mean average precision (mAP) across 32 marine species classes. This work tackles three challenges brought forward by the positive-unlabeled setting. First, does replacing the detector's built-in classification head with a separately trained classifier improve accuracy? Second, a high classifier confidence score does not guarantee the detection is a genuine match. How can we improve prediction ranking so that each detection resembles known examples of its predicted class? Third, how should we select models when the train-derived validation is unreliable? This is because the training set contains unlabeled organisms, so the detections of unannotated organisms may be interpreted as false positives against the incomplete ground truth.

Over the course of our experiments, we observed a significant collapse between local validation scores and public leaderboard performance, and after COCO URL parsing, we traced the issue to the source distribution mismatch between train and test images. This shifted our validation strategy toward external Monterey Bay Aquarium Research Institute (MBARI)-sourced evaluation data for model selection, as allowed by the competition rules. These external annotations were useful for validation, but they should not be interpreted as a fully labeled replacement for the hidden test set.

We report two closely related top submissions: the model that performed best on the public leader board and the model that performed best on the private leader board. Both use the same multi-stage foundation. Our best private-leaderboard approach is a multi-stage pipeline. First, a frozen Megalodon \cite{FathomNetMegalodon} YOLOv8x detector generates bounding box proposals using a tiled pass and a global pass merged using Non-Maximum Suppression with tile edge filtering. Second, a DINOv3 ViT-H classifier trained on the detector proposals. Lastly, the final proposal ranking is determined by a weighted geometric fusion of detector confidence and classifier confidence. This pipeline achieved a private leader board score of 0.1757 on the Kaggle evaluation set, placing us in 12th place out of 102 teams. We also evaluated a Turing Test Network (TTN) \cite{Griffin_2026_TTN}-inspired validity head as an addition reranking signal, which achieved our best public Kaggle score of 0.1864 and a private score of 0.1753. To support reproducibility, the implementation, configuration files, and experiment scripts for the full pipeline are publicly available at the DS@GT (Data Science at Georgia Tech) FathomNet CLEF 2026 Github Repo \cite{dsgt_fathomnetclef2026_code}.

\subsection{Dataset Overview}
The competition dataset is a subset of the FathomNet dataset \cite{katija2022fathomnetglobalimagedatabase}, a collection of underwater images spanning 30+ years taken using various vehicles operated by multiple institutions. Annotations follow the COCO object detection format, providing per-image bounding boxes and class labels for one of 32 marine species. 

\begin{figure}[tpb]
    \centering
    \includegraphics[width=1\linewidth]{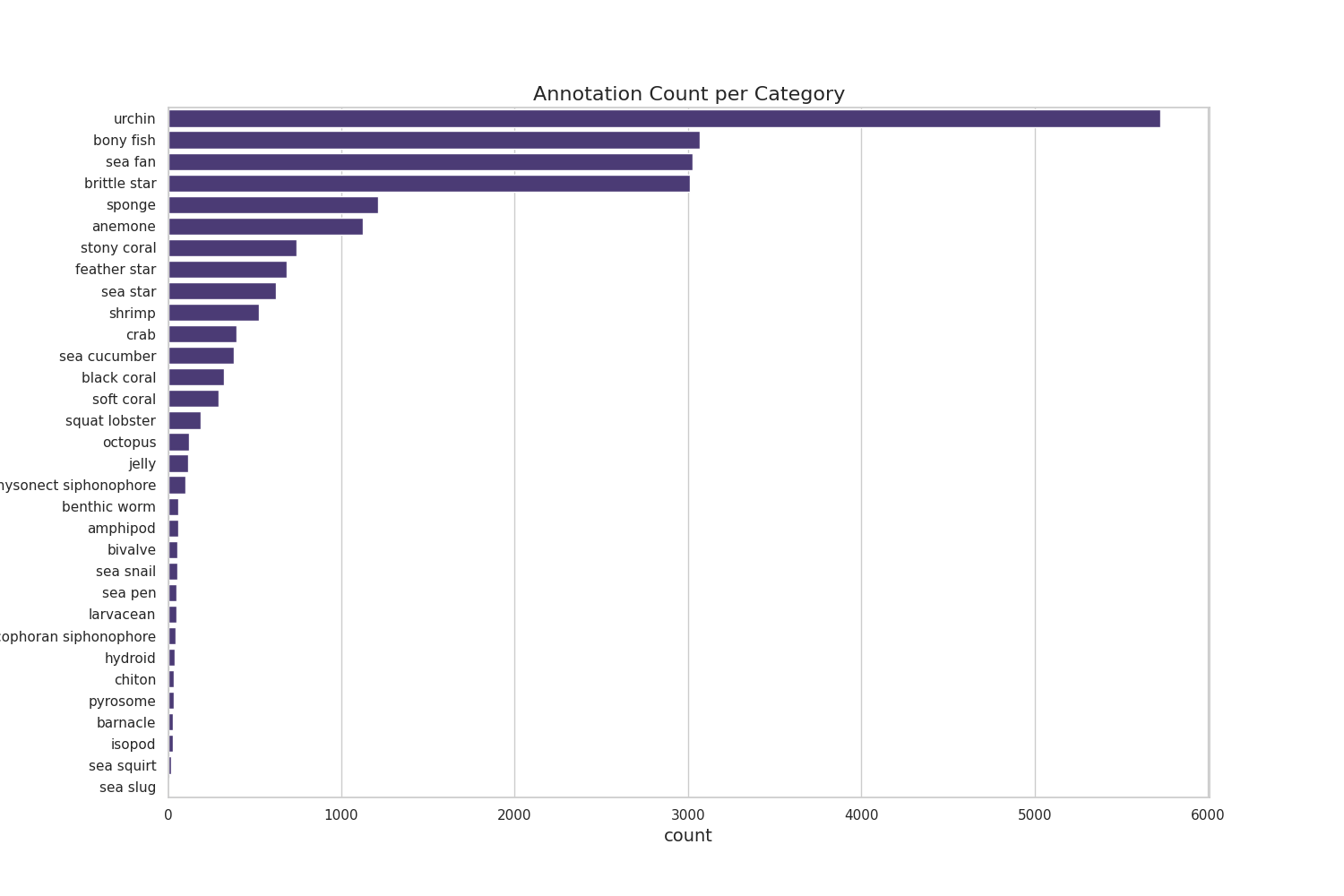}
    \caption{Distribution of classes in the training set}
    \label{fig:class-balance}
\end{figure}

\begin{figure}[tpb]
    \centering
    \includegraphics[width=0.33\linewidth]{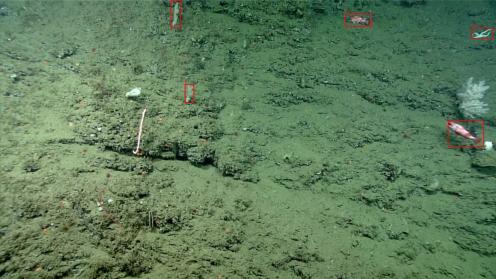}
        \includegraphics[width=0.33\linewidth]{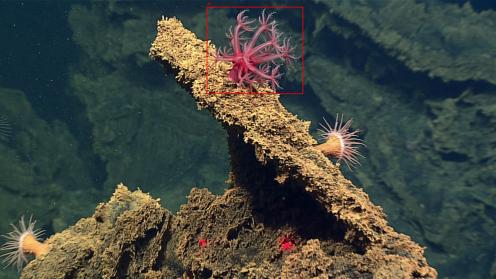}
    \includegraphics[width=0.33\linewidth]{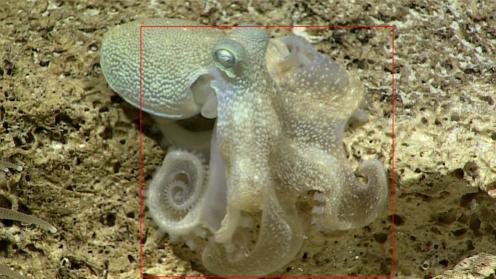}

    \caption{Three examples from the training set with their bounding boxes overlaid}
    \label{fig:train_sample}
\end{figure}

The training set contains 6,463 images sourced from National Oceanic and Atmospheric Administration (NOAA) and Schmidt Ocean Institute (SOI), captured between 2017-2023, with both labeled and unlabeled positive instances. The set exhibits a long-tailed class distribution as seen in Figure \ref{fig:class-balance}. 26$\%$ (5,723 out of 22,225) of the annotations are sea urchins, 9 classes has less than 50 instances, and the average number of annotations per image is 3.44. Bounding box dimensions are also long tailed, most boxes span less than 20$\%$ of the image width or height, though some do span the entire image. Aspect ratios are predominantly square, with a tail toward elongated 1:5 ratios. Image resolutions are split roughly 50/50 between 1920x1080 and 3840x2160. A sample of the images with their bounding boxes is shown in Figure \ref{fig:train_sample}.

The test set contains 1,425 images sourced from the Monterey Bay Aquarium Research Institute (MBARI), captured between 1999-2021, with ground-truth annotations withheld by the organizers. Image resolutions are approximately 83$\%$ 1920x1080, 13.8$\%$ 720x486, with the remainder varying. The absence of any shared institutions between train and test sets means different submersibles, expeditions, and ocean regions are represented. The out-of-distribution nature of the test set introduces a significant domain shift that compounds the positive-unlabeled challenge. 

\section{Related Work}
Previous work by the DS@GT group for PlantCLEF 2025 challenge \cite{gustineli} used a tiling-based classification strategy where images are partitioned into sub-images sized to match the vision transformer's (ViT) native input resolution. Overlapping tiles by Unel et al. \cite{Unel_2019_CVPR_Workshops} and specifically Slicing Aided Hyper Inference (SAHI) \cite{SAHI} has been used to limit detail loss for object detection. Decoupling object detection and marine species classification yielded significantly improved performance from work by Sharma et al. \cite{Sharma_2024_CVPR}, where they used MBARI underwater datasets for their work. Lastly, previous work \cite{dhlee2025fathomnet} in the Fathomnet 2025 \cite{fathomnet2025} competition passed multiscale images and full-scaled images centered on the region-of-interest (ROI) into ViT, combining object embeddings with context patch embeddings. 

\section{Methodology}
\begin{figure}[tpb]
    \centering
    \includegraphics[width=0.6\linewidth]{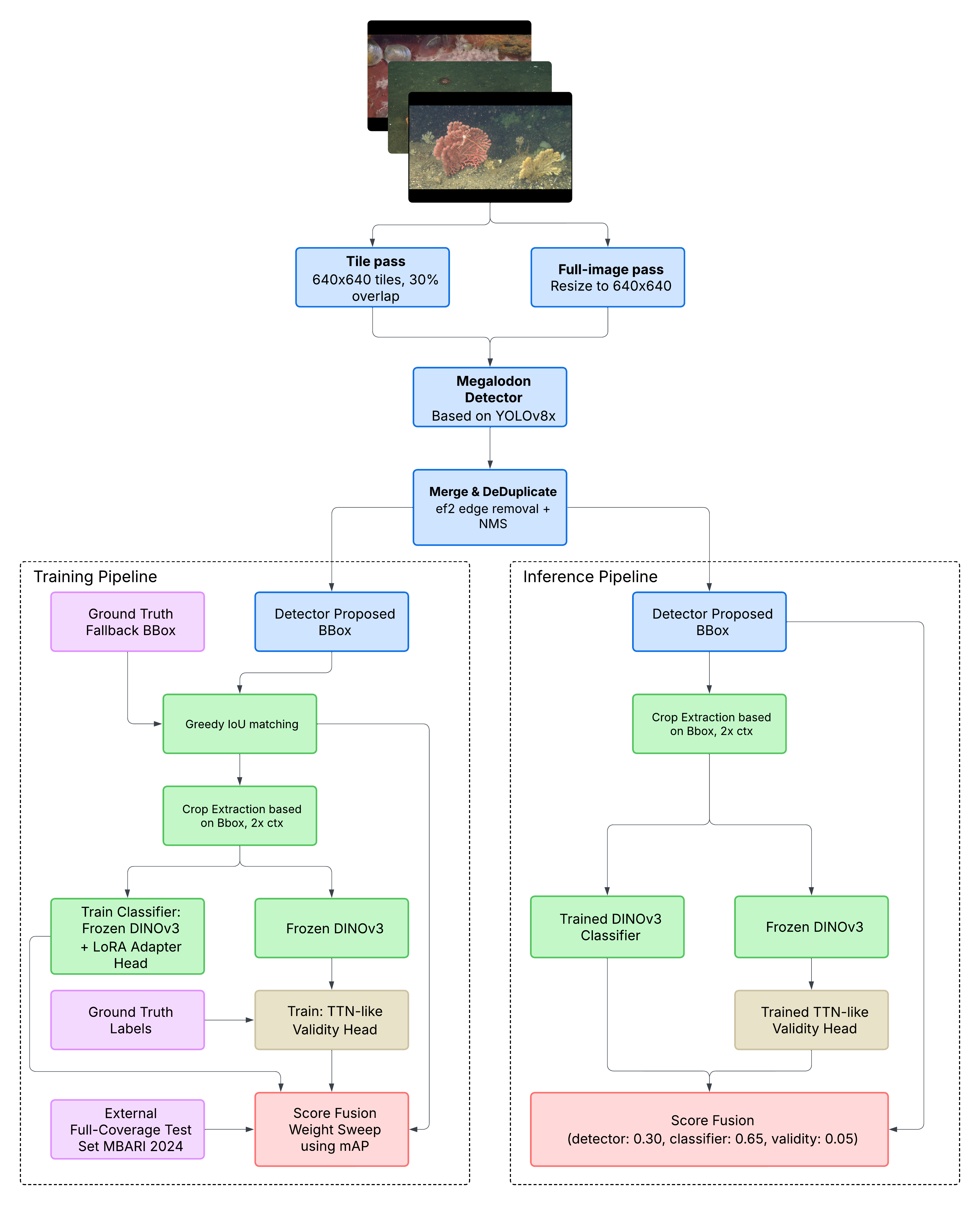}
    \caption{Pipeline of our best approach}
    \label{fig:pipeline}
\end{figure}

Our pipeline is illustrated in Figure \ref{fig:pipeline}. First, a frozen Megalodon YOLOv8x detector generates bounding box proposals via a global pass and a tiled pass, merged using Non-Maximum Supression (NMS) with tile edge filtering. Second, a DINOv3 ViT-H classifier, finetuned using Low-Rank Adaptation (LoRA) \cite{lora}, assigns species labels to each cropped proposal. Finally, detector confidence and classifier confidence are combined via weighted geometric fusion to produce the private-best ranked prediction. For the public-best submission, we add a TTN-inspired validity head as a third fusion score. 

\subsection{Decoupling Detection and Classification}
We experimented with using single end-to-end models for both detection and classification: DINOv3 ViT-L \cite{dinov3} + Faster R-CNN \cite{fasterRCNN}, DINOv3 + DINO-DETR \cite{dino-detr}, Faster R-CNN + pseudo-labels, and Swin-L \cite{swin} + DINO-DETR. We selected the DINO family as our detection framework backbone since it is a robust vision transformer which showed promising results in previous LifeCLEF competitions \cite{gustineli}\cite{miyaguchi2025dsgtanimalcleftripletlearning}, and image resolution flexibility. Swin-L was chosen as an alternative backbone for its hierarchical feature extractor. DINO-DETR was chosen as an alternative framework to Faster R-CNN since it eliminated the need for anchor boxes, thus reducing complexity. 

We observed that end-to-end detection plateaued around 0.12 mAP on the public leaderboard, which indicated to us that this approach could not overcome the positive-unlabeled setting. The unlabeled organisms act as background, which meant the model is penalized every time it correctly detects a real organism but was not annotated. This motivated a shift to a decoupled pipeline. Separating detection and classification allows a class-agnostic detector to focus on generating bounding boxes and pass context aware cropped proposals to the classifier. This meant that the classifier trains exclusively on matched ground-truth crops with known ground-truth labels, focusing on fine-grained classification without learning to suppress detection to avoid positive unlabeled penalties.  

\subsection{Detector Decisions}
We use the frozen pretrained MBARI Megalodon as our object detection proposal generator because it is from the YOLO family and it was trained to detect 1-class "object" using the full FathomNet dataset. Running the detector only on full images resized to 640x640 causes small organisms to shrink to a few pixels, making them invisible to the detector. To address this, we also generate proposals using a tiled pass that crops each image into 640x640 patches at 30$\%$ overlap. To support our decision, we used the external MBARI set to evaluate detector performance. We compared global-only, tile-only, and combined strategies and found that the combined strategy achieved the best external MBARI mAP. 

For the tile pass, organisms can end up being split across tiles which can affect the detector confidence scores. We address this by using tile edge filtering prior to merging the detections. Any tile detection with at least one bounding box edge falling within 2 pixels (ef2) of a tile edge is discarded, image boundaries do not count for edge filtering. The surviving tile detections are then concatenated with the global detections and de-duplicated using greedy NMS at an IoU threshold of 0.5. This method means the proposal boxes are preserved rather than averaging overlapping detections.

We evaluated two alternatives to the frozen stack: (1) Megalodon finetuned on tiled images with ignore-region masking at various thresholds, and (2) source-aware proposal reranking, which re-weighted detection scores based on whether a proposal originated from the global pass, the tiled pass, or appeared in both. For both evaluations, proposals were generated using the same tile and global pass, crops were expanded 2x the bounding box size to capture context, and passed to a frozen DINOv3 ViT-L classifier with log-space fusion weights of 0.5 (detector) and 1.5 (classifier).

\subsection{External MBARI Evaluation Views}

An external MBARI-sourced dataset was selected primarily as a validation resource after we observed that the provided training and test sets did not appear to be drawn from the same distribution. Because of this mismatch, performance on the training split alone was not a reliable indicator of expected leaderboard performance. In addition, Kaggle limited submissions to 10 per 24-hour period, which made it impractical to use leaderboard feedback as the main mechanism for model selection. We therefore sought an alternative evaluation signal that would better reflect out-of-distribution generalization and help distinguish genuinely improved models from runs that only performed well on the in-competition training data.

We constructed the external views from the FathomNet full-coverage test-set metadata, restricted to FathomNet database UUID records with available image metadata and bounding boxes. We then selected the subset whose image URLs matched the MBARI VARS framegrab namespace, \texttt{database.fathomnet.org/static/m3/framegrabs/}, which was the same URL pattern observed for the hidden competition test images. Raw FathomNet annotations use fine-grained marine concept names aligned with WoRMS, the World Register of Marine Species, whereas the competition evaluates only 32 broader CLEF target categories. We therefore built a manually reviewed crosswalk from each observed FathomNet/WoRMS concept to either one CLEF category or to a dropped label. For example, species- or genus-level concepts were collapsed into broader competition classes when the taxonomy matched a target class, while non-biological labels, overly broad taxa, and concepts outside the 32 target categories were excluded. Annotations whose concepts did not have an accepted mapping were omitted from the mapped evaluation annotations. The exported COCO category table used the official competition category ids and names.

This produced two COCO-formatted MBARI evaluation views, summarized in Table~\ref{tab:external-mbari-views}. The full MBARI view, \texttt{external\_mbari}, retained every MBARI-framegrab image with at least one mapped annotation. The strict view, \texttt{external\_mbari\_strict}, further retained only images for which every available FathomNet annotation mapped cleanly into one of the CLEF target classes.

\begin{table}[tpb]
    \caption{External MBARI evaluation views used for model selection and diagnostics.}
    \centering
    \begin{tabular}{p{0.26\linewidth}p{0.46\linewidth}rr}
        \toprule
        View & Selection rule & Images & Boxes \\
        \midrule
        \texttt{external\_mbari} & MBARI framegrab images with at least one annotation mapped to a CLEF target class & 1,191 & 5,575 \\
        \texttt{external\_mbari\_strict} & Subset of \texttt{external\_mbari} where every available annotation mapped to a CLEF target class & 125 & 423 \\
        \bottomrule
    \end{tabular}
    \label{tab:external-mbari-views}
\end{table}

The strict MBARI view should not be interpreted as exhaustively labeled. Rather, ``strict'' means that no available annotation on a retained image belonged to an unmapped or out-of-scope concept. Unlabeled organisms could still be present. The full MBARI view provided the larger and more stable model-selection signal, while the strict view served as a cleaner but smaller companion diagnostic for category-mapping noise. Unless otherwise stated, tables reporting external MBARI mAP use the full \texttt{external\_mbari} view.

Although the MBARI dataset could likely have been incorporated into training, we chose not to use it for model training. Our view was that doing so would be undermining the positive-unlabeled and domain shift challenges. Instead, we used it only as an external reference for validation and model comparison.

\subsection{Classifier Model Decisions}

We selected DINOv3 \cite{dinov3} as the pretrained classifier for several reasons. In preliminary experiments conducted before the competition, DINOv3 performed well on zero-shot classification tasks using a small but relevant marine life dataset \cite{marine_life_kaggle}. These early results suggested that it was a strong candidate for the classification component of the pipeline.

This choice was further supported by subsequent LoRA adapter fine-tuning experiments comparing DINOv3 and Bio-CLIP v2.5 \cite{Gu_BioCLIP_2.5_Huge_model}. When classifying the bounding box proposals generated by the detector, the DINOv3-based pipeline achieved a public leaderboard mAP of 0.1643, whereas the Bio-CLIP v2.5 pipeline underperformed significantly, yielding an mAP of 0.0599. This result was initially unexpected given that Bio-CLIP\cite{gu2025bioclip} includes FathomNet-related data in its training corpus.

We hypothesize that this performance gap stems from two structural factors:
\begin{itemize}
    \item \textbf{Downstream Proposal Misalignment:} DINOv3's pure Vision Transformer architecture proved highly robust to the variable geometry, padding, and resolutions of the detector's raw crop outputs. Conversely, the CLIP visual backbone may have been more sensitive to these framing differences, causing a domain mismatch during feature extraction.
    
    \item \textbf{Text-Space Alignment and Taxonomic Hierarchy:} As a dual-encoder multimodal model, Bio-CLIP's text encoder was heavily pretrained on precise, granular scientific names (e.g., specific genus and species). In contrast, the FathomNetCLEF 2026 dataset targets a mixture of broad common names and general species groups (e.g., ``urchin'', ``bony fish''). Forcing the text-space embeddings to map to these generalized targets likely disrupted the model's learned taxonomic hierarchy, whereas DINOv3's decoupled vision-only representation adapted seamlessly to the target label space during LoRA fine-tuning.
\end{itemize}

\subsection{TTN-Inspired Validity Reranking}
Classifier confidence alone can be overconfident on visually plausible but incorrect candidates. A high softmax score can reflect the absence of a better class rather than a genuine match, especially with the domain shift in this competition. To address this, we introduce a validity reranking step inspired by Turing Test Networks (TTN) \cite{Griffin_2026_TTN}.

Our validity stage is a 6-layer transformer encoder trained on the embeddings from the frozen DINOv3 classifier. For each candidate proposal, the model takes as a sequence of 10 reference crops from the same predicted class followed by the candidate embedding and outputs a logit that represents: does the candidate crop resemble known members of its predicted class? To train the model, we create synthetic positive and negative pairs from the ground truth labels. A positive pair is where the candidate matches the references, while a negative pair matches the references with an embedding from a different class. This avoids the need for manually labeled negatives. 

Our validity stage is different from TTN by (1) implementing our own model instead of using the frozen TTN provided by Griffin, (2) we used embeddings from our own DINOv3 classifier rather than the TTN embedding stack, and (3) we used the validity score for score fusion rather than applying a threshold that removes low-scoring proposals entirely.

\subsection{Score Fusion}
The fusion weights are selected by running inference on the external MBARI set to obtain detector, classifier (and TTN validity scores for best public approach), then sweeping over weight combinations and selecting the combination that maximizes mAP. Geometric fusion is used because it penalizes candidates that are weak in any stage of the pipeline, leaving the strongest candidates across each stage near the top of our prediction rankings. 

\subsection{Implementation Details}

\begin{itemize}
    \item \textbf{Proposal Generation \& Crop Handling:} Tiled inference was conducted with a 30\% overlap. Detector bounding box proposals were merged using Non-Maximum Suppression (NMS) with an edge-filter epsilon of 2.0. For classification, cropped proposals were expanded to 2.0x the bounding box size to capture surrounding context and subsequently resized to 384x384 pixels.
    
    \item \textbf{Classifier \& Validity Training:} The classifier backbone, a pretrained DINOv3 ViT-H+ model (\texttt{dinov3-vith16plus-pretrain-lvd1689m}), was kept frozen while the LoRA adapter weights were fine-tuned. LoRA was applied to the \texttt{q\_proj}, \texttt{k\_proj}, \texttt{v\_proj}, \texttt{o\_proj}, \texttt{up\_proj}, and \texttt{down\_proj} modules with a rank of 16, an alpha of 16, and a dropout of 0.1. Training was conducted for up to 30 epochs (with an early stopping patience of 6) using a batch size of 24, 2 gradient accumulation steps, and \texttt{bf16} precision. We used a learning rate of 0.0002 for the LoRA adapters and 0.001 for the classification head, applying a weight decay of 0.01 and a warmup fraction of 0.05. 
    
    \item \textbf{Inference \& Score Fusion:} Feature embeddings for the validity head were extracted using bfloat16 (\texttt{bf16}) precision with a batch size of 64. The TTN-inspired validity head was trained on sequences comprising 10 reference embeddings and 1 candidate embedding. Same-class pairs served as positive examples, while wrong-class pairs were used to generate synthetic negatives. Final downstream predictions were generated using an evaluation batch size of 64. The geometric fusion weights were determined via a sweep over an out-of-distribution held-out validation set of MBARI images, maximizing for external MBARI dense mAP (end-to-end detector + classifier mAP). This sweep resulted in the final applied weights: detector = 0.30, classifier = 0.65, and validity = 0.05.
\end{itemize}

\section{Results}

\subsection{Main Leaderboard (LB) Result}

Table \ref{tab:main-leaderboard-results} summarizes the two central submitted systems. Both models share the frozen Megalodon proposal stack and DINOv3 ViT-H classifier.

\begin{table}[tpb]
     \caption{Main submitted systems. Public and private scores are Kaggle mAP.}
    \centering
    \begin{tabular}{p{0.46\linewidth}ccc}
        \toprule
        System & Public LB & Private LB & External MBARI mAP \\
        \midrule
        Frozen Megalodon global+tile proposals, DINOv3 ViT-H classifier, detector/classifier fusion
            & 0.1809 & \textbf{0.1757} & 0.131830 \\
        Same system with TTN-inspired validity fusion
            & \textbf{0.1864} & 0.1753 & \textbf{0.141452} \\
        \bottomrule
    \end{tabular}
    \label{tab:main-leaderboard-results}
\end{table}

\subsection{Decoupling Detection and Classification Results}
Results for single end-to-end and our initial decoupling results are shown in Table \ref{tab:model-study}. The results show that instead of finding a one-size-fits-all backbone for both detection and classification, decoupling the two stages yields better results. The 0.1119 decoupled pipeline scored just below the best single-model result of 0.1201, which suggests that tiling is needed to realize the gains of decoupling.

\begin{table}[]
    \caption{Single End-to-End Experiment Results}
    \centering
    \begin{tabular}{lc}
        \toprule
         Architecture & Public mAP \\
         \midrule
         Single end-to-end pipeline\\
         \hline
        DINOv3 ViT-L + Faster R-CNN (frozen, 640x640 scale) & 0.0316 \\
        DINOv3 ViT-L + Faster R-CNN (unfrozen, 640x640 scale) & 0.0975\\
        DINOv3 ViT-L + DINO-DETR & -$^+$\\
        DINOv3 ViT-L + Faster R-CNN (1280/1536 multiscale, unfrozen, warmup=1000) & 0.1201\\
        Faster R-CNN + pseudo-labels (confidence threshold $\geq$ 0.5, NMS @ IoU 0.5)  & 0.1141\\
        Swin-L + DINO-DETR (2 Swin stages frozen, 1280/1536 multiscale) & 0.0738\\
        \hline
        Decoupled Pipeline\\
        \hline
        Megalodon (1280 scale, no tiling) + DINOv3 ViT-L & 0.1119\\
        Megalodon (SAHI tiling) + NMS-ef2 + DINOv3 ViT-L & \textbf{0.1581}\\
        \bottomrule
        +: validation mAP collapsed during training
    \end{tabular}
    \label{tab:model-study}
\end{table}

\subsection{Detector Decision Results}
The results from global vs tiling input for the detector experiments are shown in Table \ref{tab:tiling-decision}. Adding a tile pass to global-only improved mAP slightly, which we incorporated into the pipeline.

The results from the frozen vs finetune experiment are shown in Table \ref{tab:detector-decision}. Note that the mAP scores are for the detector stage only to isolate effects. While the finetuned Megalodon detectors outperformed the frozen Megalodon on the external MBARI validation set, their best checkpoints were at the initial epochs and the mAP declined monotonically after. In addition, the best performing finetuned Megalodon scored 0.1252 on the Kaggle public leaderboard, while the frozen Megalodon scored 0.1581. Based on this, we chose the frozen model as our detector for a simpler pipeline.

For the source-aware alternative, we evaluated these experiments using the external MBARI set on the full pipeline. Source-aware reranking of detection proposals scored 0.1158 on the external MBARI validation set, slightly less than the frozen Megalodon of 0.1231.

\begin{table}[]
    \caption{Global and Tiling Passes Experiment Results}
    \centering
    \begin{tabular}{lc}
        \toprule
         Method & External MBARI mAP (Detector only) \\
         \midrule
        Global only & 0.2823 \\
        Tile only (ef2) & 0.2454\\
        Global + tile (NMS ef2) & \textbf{0.2834}\\
        \bottomrule
    \end{tabular}
    \label{tab:tiling-decision}
\end{table}

\begin{table}[]
    \caption{Finetune vs Frozen Megalodon Experiment Results}
    \centering
    \begin{tabular}{ccc}
        \toprule
         Ignore Threshold & External MBARI mAP (Detector only) & Peak Epoch \\
         \midrule
        0.15 & 0.3086 & 1 \\
        0.20 & 0.3231 & 1\\
        0.25 & 0.3230 & 1\\
        0.35 & \textbf{0.3381} & 1\\
        0.40 & 0.3307 & 1\\
        0.45 & 0.3211 & 0\\
        \hline
        Frozen Megalodon & 0.2967 \\
        \bottomrule
    \end{tabular}
    \label{tab:detector-decision}
\end{table}

\subsection{TTN-Inspired Validity Reranking}
The TTN-inspired validity head was added to improve ranking among classifier outputs. It assigned each candidate an auxiliary class-consistency score, which was fused with detector and classifier confidence. Table \ref{tab:ttn-validity-results} shows the key ablation. The validity head improved external MBARI dense mAP by 0.009622 and improved the public leaderboard by 0.0055. However, the private leaderboard slightly favored the no-validity fusion. This was the main public/private reversal in the final experiments.

\begin{table}[tpb]
    \caption{Effect of TTN-inspired validity fusion.}
    \centering
    \begin{tabular}{p{0.42\linewidth}ccc}
        \toprule
        System & Public LB & Private LB & External MBARI mAP \\
        \midrule
        Detector/classifier fusion only & 0.1809 & \textbf{0.1757} & 0.131830 \\
        Detector/classifier/validity fusion & \textbf{0.1864} & 0.1753 & \textbf{0.141452} \\
        \bottomrule
    \end{tabular}
    \label{tab:ttn-validity-results}
\end{table}

\subsection{Chronological Experiment Summary}
Table \ref{tab:chronological-results} summarizes the experimental progression. The early experiments establish the limits of end-to-end training and pseudo labels under sparse annotations. The turning point was the source audit, which shifted model selection toward MBARI-like external evaluation. The final phase combined the frozen proposal stack, DINOv3 classifier, calibrated score fusion, and TTN-inspired reranking experiments.

\begin{table}[tpb]
    \caption{Chronological summary of experimental findings.}
    \centering
    \begin{tabular}{p{0.22\linewidth}p{0.32\linewidth}p{0.34\linewidth}}
        \toprule
        Period & Main result & Interpretation \\
        \midrule
        Initial approach & End-to-end FRCNN/DETR/Swin and pseudo-label lines plateaued below the final system & Sparse labels and train-derived validation were not enough for reliable model selection \\
        Turning point & COCO URL audit showed train/test source mismatch & Model selection shifted toward MBARI-style external evaluation and leaderboard confirmation \\
        OOD-first rethink & Frozen Megalodon proposal stack plus DINOv3 classifier became the strongest private line & Proposal quality, classifier calibration, and source-aware validation mattered more than detector fine-tuning alone \\
        TTN-inspired reranking & Improved public leaderboard and external MBARI, but not private leaderboard & Useful ranking signal, but not robust enough to become the private-best component \\
        \bottomrule
    \end{tabular}
    \label{tab:chronological-results}
\end{table}


\section{Discussion}

\subsection{Public and Private Leaderboards Rewarded Slightly Different Rankings}
The final model comparison is best understood as a split between two useful rankings, not as a clean win or loss for the TTN-inspired validity head. The validity model improved the full external MBARI score from 0.131830 to 0.141452 and improved the public leaderboard from 0.1809 to 0.1864. However, the private leaderboard slightly preferred the detector/classifier-only fusion, 0.1757 versus 0.1753. This difference is small, but it is important methodologically: a ranking signal can be real on one split and still change the ordering of borderline private examples enough to lose on the final split. Table~\ref{tab:discussion-public-private} summarizes this public/private reversal.

\begin{table}[tpb]
    \caption{Public/private reversal for the TTN-inspired validity signal.}
    \centering
    \begin{tabular}{p{0.25\linewidth}p{0.2\linewidth}p{0.1\linewidth}p{0.1\linewidth}p{0.2\linewidth}}
        \toprule
        Model & Ext. MBARI mAP & Public LB & Private LB & Interpretation \\
        \midrule
        Detector/classifier fusion & 0.131830 & 0.1809 & \textbf{0.1757} & best private ranking \\
        Detector/classifier/validity fusion & \textbf{0.141452} & \textbf{0.1864} & 0.1753 & best public/offline ranking \\
        \bottomrule
    \end{tabular}
    \label{tab:discussion-public-private}
\end{table}

We therefore report both systems: the TTN-inspired model as the best public and external-MBARI system, and the detector/classifier fusion as the best private-leaderboard system.

\subsection{Why the TTN-Inspired Validity Head Helped, but Only as a Light Prior}
The TTN-inspired head was trained to provide a class-conditional validity score for a candidate crop. Its strongest setting used a very small validity exponent in the final weighted geometric mean: detector 0.30, classifier 0.65, and validity 0.05. This is a useful diagnostic result. The validity score was not strong enough to replace detector or classifier confidence, but it was useful as a small reranking nudge on the external MBARI distribution. Table~\ref{tab:discussion-validity-sweep} shows the main evidence from the validity fusion sweep.

\begin{table}[tpb]
    \caption{Validity-head evidence from the fusion sweep.}
    \centering
    \begin{tabular}{p{0.38\linewidth}cp{0.38\linewidth}}
        \toprule
        Scoring line & Ext. MBARI mAP & Discussion point \\
        \midrule
        Detector/classifier baseline, weights 1.00/2.00/0.00 & 0.131830 & strong base ranking \\
        Tuned detector/classifier/validity, weights 0.30/0.65/0.05 & \textbf{0.141452} & best external MBARI point \\
        Validity-only control & 0.055051 & validity is not a standalone confidence \\
        Source-aware proposal family plus tuned validity & 0.128146 & weaker TTN partner than the baseline proposal family \\
        \bottomrule
    \end{tabular}
    \label{tab:discussion-validity-sweep}
\end{table}

This supports two claims. First, reference-conditioned validity captured a real class-consistency signal: otherwise it would not have improved external MBARI and public leaderboard scores. Second, the signal was not robust enough to dominate the detector/classifier ranking. A plausible explanation is that the validity head learned source- or class-frequency structure that aligned with the public split and external MBARI set, but slightly misranked some private examples. In this setting, validity was most effective as a weak prior over an already strong detector/classifier ranking rather than as a standalone confidence estimate.

\subsection{Why We Kept the Frozen Megalodon Proposal Stack}
The fine-tuned Megalodon experiments are a useful cautionary example because their detector-only results initially looked promising. Ignore-region fine-tuning improved detector AP@.5 and production-point precision relative to a baseline fine-tune. However, the improvement did not transfer to the end-to-end Kaggle pipeline, and later sweeps showed that the fine-tuning gain was transient. Table~\ref{tab:discussion-finetuned-megalodon} summarizes the evidence for this decision.

\begin{table}[tpb]
    \caption{Why fine-tuned Megalodon was not promoted.}
    \centering
    \begin{tabular}{p{0.35\linewidth}p{0.24\linewidth}p{0.30\linewidth}}
        \toprule
        Evidence & Result & Interpretation \\
        \midrule
        Ignore-loss detector-only AP@.5 & 0.4175 vs 0.3386 baseline fine-tune & fine-tuning can improve train-adjacent detector AP \\
        Production-point precision & 0.293 vs 0.164 baseline fine-tune & not a simple precision-drop story \\
        Kaggle transfer & 0.1252 fine-tuned vs 0.1581 frozen proposal line & detector-only AP did not transfer \\
        Callback sweep & vanilla 0.2967 fc/mAP (full-coverage MBARI evaluation splits); fine-tuned peaks near 0.31--0.34 at epoch 0--1 then declines & fine-tuning gain was unstable \\
        \bottomrule
    \end{tabular}
    \label{tab:discussion-finetuned-megalodon}
\end{table}

The main lesson is that detector selection had to be judged inside the full proposal, crop-classification, and ranking pipeline. Fine-tuning improved some objectness and calibration measures, but it also changed box geometry and score distributions in ways that hurt dense mAP after classification. The experiments also identified box geometry as a bottleneck: weighted box fusion and tile-boundary fragments could move boxes away from the ground truth at high IoU thresholds. This explains why tile-edge filtering and non-averaging NMS became practical decisions, even though they look like small post-processing details.

The fine-tuning results therefore changed the interpretation of detector adaptation. Fine-tuning was not rejected because it failed to learn from the competition data. It was rejected because its gains were unstable and did not survive the full proposal-classification-ranking pipeline..

\subsection{Why Source-Aware Proposals Were Informative but Not Final}
Source-aware proposal scoring was motivated by a real observation: proposal origin carried information. Boxes supported by both global and tiled passes, or boxes arising from certain source patterns, behaved differently from isolated tile-only fragments. The source-aware line improved several detector-side ranking diagnostics and performed well on the strict MBARI slice, where all retained annotations mapped into the 32 CLEF classes. The problem was that it did not beat the simpler global-plus-tiled edge-filtered proposal family on the larger full MBARI downstream evaluation with the classifier in the loop. Table~\ref{tab:discussion-source-aware} shows this strict-versus-full MBARI tradeoff.

\begin{table}[tpb]
    \caption{Source-aware downstream results with the frozen classifier stack.}
    \centering
    \begin{tabular}{lcc}
        \toprule
        Proposal family & Strict MBARI mAP & Full external MBARI mAP \\
        \midrule
        Global+tile edge-filtered proposals & 0.25790 & \textbf{0.12309} \\
        Source-aware v2 & \textbf{0.27669} & 0.11583 \\
        Source-aware v3 & 0.27661 & 0.11565 \\
        Raw source-aware v3 family & 0.25974 & 0.10965 \\
        \bottomrule
    \end{tabular}
    \label{tab:discussion-source-aware}
\end{table}

This strict/full disagreement kept source-aware scoring as an informative diagnostic rather than the final proposal family: promotion required improving the larger full-MBARI downstream pipeline, not only the cleaner strict slice.

\subsection{Why Aggressive Filtering and a Negative Class Were Risky}
One tempting response to false positives is to add a background or negative class, or to filter out low-confidence and high-risk predictions. Visual audits showed that many predictions counted as false positives against local annotations were real organisms that were simply unlabeled or outside the 32 competition classes. Post-hoc CSV filtering on a 0.1252 public leaderboard baseline consistently reduced Kaggle score, as shown in Table~\ref{tab:discussion-filtering}.

\begin{table}[tpb]
    \caption{Post-hoc filtering degraded a Kaggle-bound submission.}
    \centering
    \begin{tabular}{p{0.44\linewidth}cc}
        \toprule
        Variant & Rows kept & Public LB \\
        \midrule
        No manipulation baseline & 53,114 & \textbf{0.1252} \\
        Top-50 predictions per image & 47,748 & 0.1241 \\
        Per-class classifier floor & 40,893 & 0.1210 \\
        Classifier floor 0.5 & 23,217 & 0.1183 \\
        Combined detector/classifier floor 0.3 & 4,725 & 0.0834 \\
        Combined detector/classifier floor 0.4 & 3,510 & 0.0657 \\
        \bottomrule
    \end{tabular}
    \label{tab:discussion-filtering}
\end{table}

This does not mean a negative class is useless. A manually curated negative class could still help reject cables, equipment, texture, or non-target organisms if it were built carefully. The key  risk is that an overly broad negative class would teach the model to suppress true organisms that are absent from the sparse labels. For this competition's top-100 mAP, preserving long-tail candidate recall and improving ranking was safer than aggressively deleting uncertain detections.

\subsection{Pseudo-Label and Student-Teacher Attempts}
Positive-unlabeled detection also made pseudo-labeling and student-teacher training natural directions. If the detector could identify unlabeled organisms with sufficient reliability, those detections could be used either as additional positive examples or as teacher targets for a student model. Our experiments supported the motivation for this approach, but they also showed that the teacher signal was not strong enough to promote into the final system. Table~\ref{tab:discussion-pseudo-label-student-teacher} summarizes the main attempts.

\begin{table}[tpb]
    \caption{Pseudo-label and student-teacher attempts.}
    \centering
    \small
    \setlength{\tabcolsep}{4pt}
    \begin{tabular}{@{}p{0.24\linewidth}p{0.68\linewidth}@{}}
        \toprule
        Attempt & Main finding \\
        \midrule
        Offline self-training & A Faster R-CNN teacher increased validation mAP by roughly 30\% and improved recall, but public LB fell from 0.1201 to 0.1141 because the student copied noisy pseudo-label box geometry. \\
        Localization-only student-teacher & A vanilla Megalodon teacher with pseudo-label gates and reduced pseudo-label box loss improved clean fc-eval mAP (full-coverage MBARI evaluation splits) from 0.2802 to 0.4673, but  public LB reached only 0.1444. \\
        Frozen upstream TTNd pruning & This was the true TTNd application, using the frozen upstream TTNd checkpoint and class-conditioned prune logits rather than our locally trained TTN-inspired validity head. Raw TTNd logits contained useful ranking signal, but hard pruning was too blunt and the best soft rank-feature sweep reached 0.134705 MBARI mAP, below the local TTN-inspired reference. \\
        Classifier-side target adaptation & An EMA teacher with TTN-gated target crops produced a small MBARI mAP lift, but AR@100 dropped, so the run remained diagnostic rather than a final candidate. \\
        \bottomrule
    \end{tabular}
    \label{tab:discussion-pseudo-label-student-teacher}
\end{table}

The first pseudo-label experiment showed the core tradeoff clearly. The student found more objects and improved local validation metrics, but its boxes became less precise at stricter IoU thresholds. That is damaging under COCO-style mAP, where loose boxes can convert plausible detections into false positives. Later student-teacher training used a stronger Megalodon teacher, close-up image gating, size gates, and reduced pseudo-label box loss. This was a better formulation: it decisively improved the clean fc-eval subset (full-coverage MBARI evaluation splits) and maintained mean IoU locally. However, the leaderboard result still trailed the frozen Megalodon proposal family, showing that the local clean subset was not distribution-honest enough to select this line.

The frozen TTNd pruning attempt should be read separately from the final TTN-inspired validity head. The final validity head was trained locally and used as a light ranking feature. The TTNd pruning experiments instead used the upstream TTNd model in the intended class-conditioned form: human-labeled references, frozen TTNd weights, and prune logits for pseudo-label filtering. That more literal TTN application produced a meaningful reject-logit distribution but threshold-based pruning removed too many potential true positives and did not become part of the final model.

These experiments argue against treating pseudo-labeling as a simple data augmentation step. The method has merit for this problem because unlabeled organisms are abundant, but the teacher must be accurate, calibrated, and robust under the target distribution. Otherwise, pseudo-labels reinforce the teacher's systematic localization errors, false positives, and score biases. In our final model, it was safer to keep the frozen detector proposal family and improve ranking downstream than to train a student on teacher predictions that were not yet reliable enough.

\subsection{Why Clean Crop Classification Beat Heavier Augmentation}
The classifier results also point to a broader lesson: the main crop-classification problem was not solved by making the image distribution artificially harder. A true no-augmentation baseline beat the earlier light-augmentation control, and the out-of-distribution (OOD)-targeted augmentation variants underperformed on external MBARI even when some independent and identically distributed (IID) metrics looked strong. Table~\ref{tab:discussion-classifier-augmentation} summarizes these classifier results.

\begin{table}[tpb]
    \caption{Crop-classifier augmentation and context evidence.}
    \centering
    \begin{tabular}{p{0.34\linewidth}cc}
        \toprule
        Classifier setting & External MBARI mAP & IID validation mAP \\
        \midrule
        No augmentation, context 2.0 & \textbf{0.5545} & 0.9111 \\
        No augmentation, context 2.5 & 0.5235 & 0.8930 \\
        Legacy light augmentation & 0.5191 & 0.8813 \\
        OOD-targeted augmentation v1 & 0.5068 & 0.9060 \\
        OOD-targeted augmentation v2 & 0.5063 & 0.9144 \\
        OOD-targeted augmentation v3 & 0.5032 & 0.8885 \\
        \bottomrule
    \end{tabular}
    \label{tab:discussion-classifier-augmentation}
\end{table}

A likely explanation is that the species classifier needed clean morphological cues: body outline, texture, appendages, transparency, and local substrate context. Augmentations such as blur, color shifts, or noise may make the model more invariant in an IID split while corrupting precisely the cues needed for fine-grained marine classification. The context sweep also suggests that moderate context helped: crops that were too tight lost useful surrounding information, while too much context risked admitting nearby organisms or background patterns.

\subsection{Limitations}
Several limitations remain. External MBARI was our best offline proxy for hidden-test behavior, but it was still imperfect: the public/private reversal shows that even a domain-matched proxy did not exactly match the final private split, and strict versus full MBARI carried different annotation-noise tradeoffs. Individual ablations should therefore be interpreted as evidence within the full detector/classifier/ranking pipeline rather than as standalone wins. The competition timeline also limited exhaustive retuning, especially because proposal changes required regenerating detections, rebuilding crop manifests, rerunning classification, recomputing fusion scores, and repeating downstream evaluation.

The evaluation setup also limits how strongly we can interpret individual ablation results. Public leaderboard improvements did not always predict private leaderboard improvements, and detector-only metrics did not always predict end-to-end dense mAP after classification. These discrepancies were central to the project: they forced model selection to depend on several imperfect signals rather than a single validation number. The final system is therefore best viewed as the result of conservative evidence aggregation across external MBARI, leaderboard, and targeted diagnostic experiments.

\section{Future Work}

\subsection{Develop a process for generating additional training images}
One possible direction is to develop a pipeline for synthetic data generation using methods such as GANs \cite{GAN}, or simpler compositional approaches such as placing known valid bounding box targets onto artificial or curated background images. This would increase the amount and diversity of available training data, and may improve training performance for Megalodon or other detectors fine-tuning or LoRA adapter training. Such a process could also be useful in settings where labeled examples are limited.

\subsection{Develop a student-teacher approach for Positive-Unlabeled training}
The pseudo-label and student-teacher results in Table~\ref{tab:discussion-pseudo-label-student-teacher} suggest that
Positive-Unlabeled (PU) training remains promising, but only if the teacher is substantially more reliable than the models available during this work. Future work should combine a stronger teacher with distribution-honest validation, adaptive pseudo-label thresholds, agreement-based filtering, and explicit control over pseudo-label box-regression weight. Under those conditions, student-teacher training could make it possible to use the many unlabeled organisms in the training imagery without teaching the detector to reproduce weak teacher boxes or target-specific false positives.

\subsection{Develop calibrated filtering and ranking under sparse labels}
The filtering experiments suggest that future work should treat low-confidence detections as ranking candidates rather than simply deleting them. Hard confidence thresholds, broad negative classes, and hard TTN-style pruning all risk removing true organisms that are unlabeled, rare, or outside the available annotation set. A more promising direction is to learn calibrated soft ranking features from detector confidence, classifier confidence, proposal provenance, class-conditioned validity, and agreement between teachers. If a negative class is used, it should be narrowly curated around clear non-organism artifacts such as equipment, cables, and background texture rather than constructed from all unmatched proposals. Future promotion criteria should also track top-100 recall or AR@100 alongside mAP, so improvements in precision do not come from suppressing useful long-tail candidates.

\section{Conclusions}

This work presents a practical multi-stage approach for FathomNetCLEF 2026, where the central challenge was not only detecting and classifying marine organisms, but doing so under sparse labels and a strong train-test source shift. The final system separated object localization from species classification: a frozen Megalodon detector generated class-agnostic proposals, tiled inference recovered small organisms, tile-edge filtering removed partial tile artifacts, and a DINOv3 ViT-H classifier provided fine-grained species predictions on contextual proposal crops. Weighted geometric fusion then produced the final ranking used for dense mAP evaluation.

The experimental progression shows why this conservative design was selected. End-to-end detector/classifier models plateaued below the final system, and detector fine-tuning produced gains that were either transient or did not transfer through the full proposal-classification-ranking pipeline. Source-aware proposal scoring and TTN-inspired validity reranking both exposed useful signals, but only the latter improved public and external MBARI performance, and even then it was most effective as a small ranking prior rather than a replacement for detector or classifier confidence. The private leaderboard ultimately favored the simpler detector/classifier fusion, while the TTN-inspired variant produced the best public score and best external MBARI score.

The broader lesson is that positive-unlabeled marine imagery makes naive precision-oriented decisions risky. Many apparent false positives are likely unlabeled organisms or out-of-scope taxa, so aggressive score filtering, negative-class construction, and hard pseudo-label pruning can remove useful low-ranked true positives. Pseudo-label and student-teacher methods remain promising, especially because unlabeled organisms are common in the imagery, but our attempts showed that the teacher signal must be substantially stronger and better calibrated before it can improve the final ranking. Future work should therefore focus on distribution-honest validation, stronger teacher models, calibrated pseudo-label admission, and classifier training on realistic detector-produced crops.

Overall, our best-performing submission was not the result of a single model change, but of aligning proposal generation, crop classification, score fusion, and validation strategy with the competition's sparse-label and source-shift structure. The resulting private-best model achieved 0.1757 Kaggle mAP and ranked 12th out of 102 teams, while the public-best TTN-inspired variant demonstrated that reference-conditioned validity can provide useful ranking signal when applied as a light auxiliary prior.

\section*{Acknowledgments}

We thank the Data Science at Georgia Tech (DS@GT) CLEF competition group for their support.
This research was supported in part through research cyberinfrastructure resources and services provided by the Partnership for an Advanced Computing Environment (PACE) at the Georgia Institute of Technology, Atlanta, Georgia, USA \cite{PACE}. 

\section*{Declaration on Generative AI}
 During the preparation of this work, the authors used Claude in order to: draft content, grammar and spelling check. After using these tool(s)/service(s), the authors reviewed and edited the content as needed and takes full responsibility for the publication’s content. 

\bibliography{main}
\end{document}